# A Hybrid Semantic-Geometric Approach for Clutter-Resistant Floorplan Generation from Building Point Clouds

Seongyong Kim[1], Yosuke Yajima[2], Jisoo Park[1], Jingdao Chen[3], and Yong K. Cho[1*]

[1] *Civil and Environmental Engineering, Georgia Institute of Technology, 790 Atlantic Dr. NW, Atlanta, GA 30332, USA.* Email: skim3310@gatech.edu, jpark711@gatech.edu, yong.cho@ce.gatech.edu (*corresponding author)
[2] *Institute for Robotics and Intelligent Machines, Georgia Institute of Technology, 801 Atlantic Dr. NW, Atlanta, GA 30332, USA.* Email: yyajima@gatech.edu
[3] *Computer Science and Engineering, Mississippi State University, Mississippi State, MS 39762, USA.* E-mail: chenjingdao@cse.msstate.edu

**Abstract:** Building Information Modeling (BIM) technology is a key component of modern construction engineering and project management workflows. As-is BIM models that represent the spatial reality of a project site can offer crucial information to stakeholders for construction progress monitoring, error checking, and building maintenance purposes. Geometric methods for automatically converting raw scan data into BIM models (Scan-to-BIM) often fail to make use of higher-level semantic information in the data. Whereas, semantic segmentation methods only output labels at the point level without creating object level models that is necessary for BIM. To address these issues, this research proposes a hybrid semantic-geometric approach for clutter-resistant floorplan generation from laser-scanned building point clouds. The input point clouds are first pre-processed by normalizing the coordinate system and removing outliers. Then, a semantic segmentation network based on PointNet++ is used to label each point as ceiling, floor, wall, door, stair, and clutter. The clutter points are removed whereas the wall, door, and stair points are used for 2D floorplan generation. A region-growing segmentation algorithm paired with geometric reasoning rules is applied to group the points together into individual building elements. Finally, a 2-fold Random Sample Consensus (RANSAC) algorithm is applied to parameterize the building elements into 2D lines which are used to create the output floorplan. The proposed method is evaluated using the metrics of precision, recall, Intersection-over-Union (IOU), Betti error, and warping error.

## 1. INTRODUCTION

In Architecture, Construction, and Engineering community, Building Information Modeling (BIM) technology is a key component of modern construction engineering and project management workflows. As-is BIM models that represent the spatial reality of a project site can offer crucial information to stakeholders for construction progress monitoring, error checking, and building maintenance purposes. However, as-planned BIM models may not always be available, and for the important use case of as-built BIM that represents the actual site conditions, the approach of building the model using 3D scan data have recently gained attention. Especially, the 1st Scan-to-BIM challenge is recently hosted by the 2021 CV4AEC workshop and provide a publicly-available building point cloud dataset to reconstruct floor plans or BIM models (*https://cv4aec.github.io/*). This paper will introduce our method, which placed 2nd in the floor plan reconstruction challenge.

Geometric methods for automatically converting raw scan data into BIM models (Scan-to-BIM) often fail to make use of higher-level semantic information in the data. Whereas, semantic segmentation methods only output labels at the point level without creating object level models that is necessary for BIM. Existing methods for Scan-to-BIM are also vulnerable to occlusion, clutter, and high variability in the input point clouds. Furthermore, compared to previous public datasets, the CV4AEC dataset provides an unstructured large-scale reality dataset with significant noise and occlusion. To address these issues, this research proposes a hybrid semantic-geometric framework for clutter-resistant floorplan generation from laser-scanned building point clouds (Figure 1). The input point clouds are first pre-processed by normalizing the coordinate system and removing outliers. Then, a semantic segmentation network based on PointNet++ is used to label each point as ceiling, floor, wall, door, stair, and clutter. The clutter points are removed whereas the wall, door, and stair points are used for 2D floorplan generation. A region-growing segmentation algorithm paired with geometric reasoning rules is applied to group the points together into individual building elements. Finally, a 2-fold Random Sample Consensus (RANSAC) algorithm is applied to parameterize the building elements into 2D lines which are used to create the output floorplan.

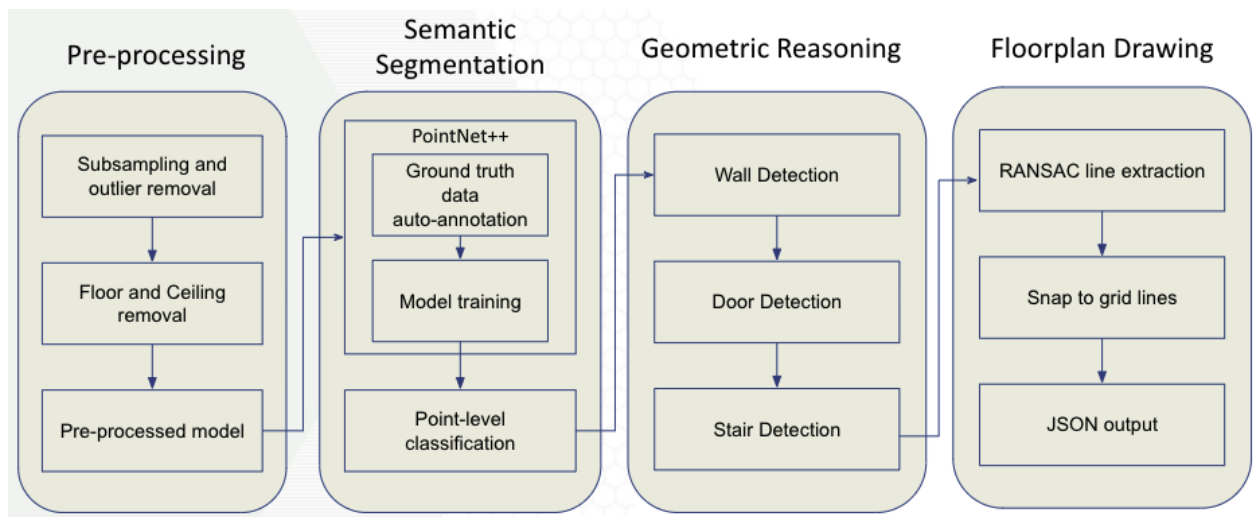

**Figure 1.** The overall framework

## 2. RELATED WORK

This section provides an overview of the related literature in the context of both building point cloud datasets and algorithms to reconstruct semantic objects from raw point clouds. As the quality and traits of the building point clouds directly affects the direction of the algorithm development, and since the CV4AEC dataset used in this paper has several differences from other commonly-used datasets, we summarize the characteristics of different building point cloud datasets in terms of object class of interest and target building elements in Table 1. Individual datasets contain different object classes, but we assign them into the following categories in Table 1: Core structural element refers to the key building elements of a wall, floor, ceiling, and door while structural element includes all of them, to which other static building parts can be added. Indoor object refers to moving or temporary objects such as furniture. Compared to others, the CV4AEC dataset we targeted has much greater data volume and contains diverse types of building areas with clutter, thus causing difficulty for the existing method to fully exploit the features of the whole datasets.

**Table 1.** Building Point Cloud Datasets

| Dataset | Description | Objects | Building Elements |
|---|---|---|---|
| S3DIS [1] | 6 large-scale building areas | 13 object classes | Structure elements, Indoor objects |
| Xue et al.'s [2] | Scanned furnitures via a smart phone | 6 furniture classes | Indoor objects |
| Han et al.'s [3] | Meeting room, church, office, 3 buildings | Wall, Ceiling, Floor, Cylinder, Room | Core structure elements |
| Zeng et al.'s [4] | Buildings under construction, Church, Office | Wall, Window, Door, Column, Roof | Structure elements (manually expandable) |
| Raamac Lab dataset [5] | Offices, Conference rooms, Storage | Wall, Ceiling, Floor, Column, Beam, Pipe | Sructure elements |
| **CV4AEC** | 91 large-scale building areas **with clutter** | Wall, Floor, Ceiling, Door, Room, Stairs | Core structure elements, **Topological elements** |

Table 2 briefly summarizes the related work from the aspects of classifier type, end results, and evaluation metric. Table 2 also demonstrates how our hybrid approach of semantic and geometric reasoning on CV4ACE differs in those points. Point-level refers to a classification result for each scanned point, while object-level refers to a classification result for building objects which is processed from clustered points. Other Scan-to-BIM related research based on learning algorithms exist [6]–[8], but most of them aim at predicting point-level classes, rather than generating building models. Beyond simply classifying the points with semantic information, we proposed the method to assemble the point chunks and fabricate them into semantically meaningful building objects, including topological relationships.

**Table 2.** Comparison of different Scan-to-BIM methods

| Method | Classifier | End results | Evaluation metric | Dependency |
|---|---|---|---|---|
| [1] | Geometric | Points with semantic info. | Point-level semantic eval., and room parsing | **Open source** |
| [2] | Geometric | Points clustered in object unit | **Object-level** eval. | Autodesk Revit |
| [3] | Geometric | LoD 2 model in CityGML | Point-level structure similarity | |
| [9] | PointNet/ DGCNN | BIM | Point-level semantic eval. | Autodesk Revit, Trimble, FME |
| **Ours** | PointNet++ & Geometric | Floorplan | **Object-level** semantic eval, **topological** eval. | **Open source** |

## 3. SEMANTIC SEGMENTATION

### 3.1. Pseudo-annotation

One of the biggest barriers to utilizing a volumetric building point cloud dataset is the demanding annotation task due to the large number of points and the difficulty of distinguishing 3D objects of interest from clutter. In a situation where it is difficult to obtain labeled data, we propose a pseudo-annotating method using incomplete existing 2D or 3D drawings.

In the CV4AEC challenge, the training data that is given is only a set of 2D drawings showing polylines of certain object classes such as a wall, door, and stair. To convert this data into a format that can be used to train 3D semantic segmentation models, we translated those polylines into 3D point clouds and mapped object classes into individual points with offsets. First, based on the histogram pattern in the z-axis, the floor height can adaptively determined, through which the floor and ceiling classes can be obtained (Figure 2). The wall, door, and stair points are then annotated based on the projected 2D drawings. Finally, the rest of the points are designated as the clutter class. Through this pseudo-annotation process, the point-level annotations can be obtained, albeit with imperfect accuracy.

In general, as-is scan data is likely to show a discrepency with the as-planned model because of the indoor temporary objects and clutter. For this reason, our method does not fully depend on semantic segmentation results, and supplements the results using geometric reasoning.

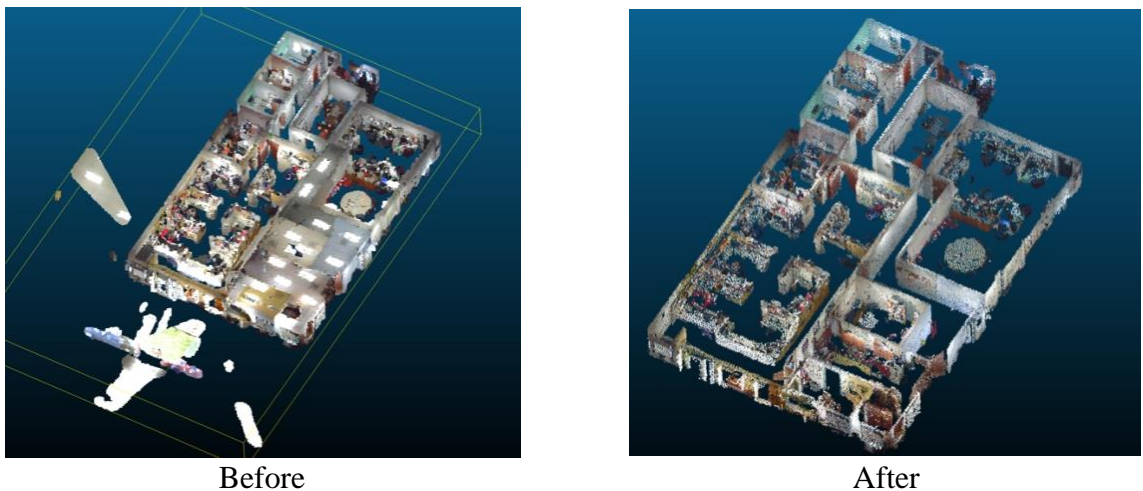

Before                              After

**Figure 2.** Pre-processing of removing noise, identifying the ceiling and floor points

### 3.2. 3D deep neural network

In training the 3D deep learning classifier, we adopt the PointNet++ architecture [10], one of the most widely-used deep networks for point cloud semantic segmentation. We perform point-level classification using the target classes of the core structure elements and clutter. On the other hand, the "stair" class is only detected based on geometric reasoning because of class imbalance in the training data. One advantage of applying semantic segmentation is having the ability to remove clutter, thus enabling post-processing of geometric reasoning to gain higher confidence in detecting building elements with clear boundaries.

## 4. GEOMETRIC REASONING

## 4.1. Wall and door detection

Starting from a extracted point cloud of only wall points, obtained from the semantic segmentation results, we generated wall and door instances to be used for reconstructing the floor plan. As the wall point set already excludes most noise and clutter, we are able to use the assumption that walls consist of smooth planer surfaces (Figure 3). First, we specified wall candidates by extracting only the points that have normal vectors that are perpendicular to the floor. The wall candidates are segmented into individual wall objects using a region growing algorithm, in which neighboring points with similar normal vectors are grouped together to form wall instances. Finally, as a post-processing step, parametric filtering based on cluster size and wall height was performed to remove outliers and increase reliability.

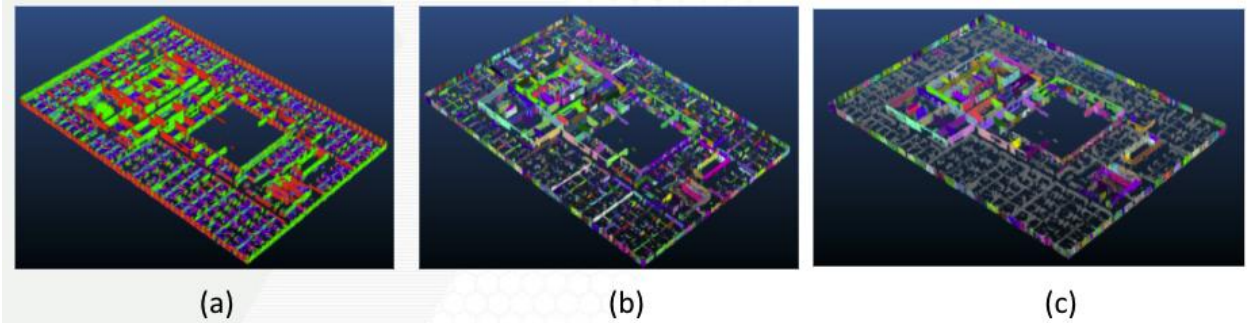

**Figure 3.** Wall detection stages: (a) normal estimation, (b) region growing, and (c) parametric filtering.

Door detection is challenging because the CV4AEC challenge criterion specifies to detect doors regardless of whether they are open, closed, or even missing from the point cloud. In this situation, we perform door detection by measuring the empty space in the wall objects (Figure 4). A vertical histogram is applied to determine square empty spaces in the plane of the wall that do not contain any points and would thus indicate the presence of a doorway. Next, parametric filtering is applied to filter out false detections based on door size and aspect ratio.

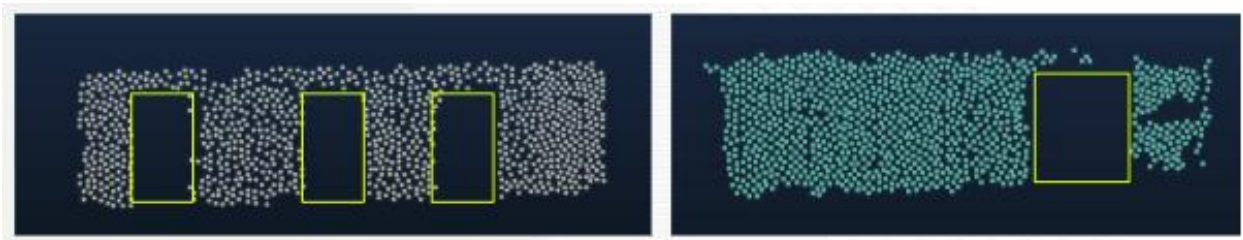

**Figure 4.** Door detection from the individual wall object

## 4.2. Stairway detection

To perform stairway detection, we applied the approach of searching for geometrical areas that resembles risers and treads that appear a row [11]. In this way, an area detected as a specific shape is regarded as a stairway. In addition, we tuned the algorithm to normalize point coordinates through adaptively moving the center coordinate, and removed outliers based on the sequence of risers and treads.

### 4.3. Floorplan drawing

The output floorplan is generated using the detected objects of walls, doors, and stairs, according to the format of the CV4AEC challenge. The reconstructed floorplan basically consists of a 2-fold polyline representing inner and outer surface of each object. All the detected objects are individually segmented on a per-room basis. To obtain 2D lines from the individually detected building objects from the previous section, we applied a 2-fold RANSAC algorithm to the top view of the 3D points cloud, thus projecting point objects into sets of 2D polylines. For stairway objects, each riser and tread is represented as a line in the floorplan.

## 5. EXPERIMENTS

### 5.1. Dataset

This research utilizes the CV4AEC dataset, consisting of 91 large-scale building point clouds. Training, validation, and test set were divided into 49, 21, and 21 point clouds, respectively, while preserving a even distribution of building types. We utilized the validation set to avoid over-fitting in deep learning section and find an optimistic parameters for parametric filtering

The dataset shows diverse properties in building type and appearance. The building types range from offices to parking lots to rooftops, thus complicating the process of parametric geometric reasoning. Besides, the type of raw data available is different per building; some point clouds do not have color information and some do not have floor and ceiling surface points (Figure 5). More importantly, the dataset is obtained from real laser-scanning, which leads to the point clouds having a significant amount of occlusion, missing spots, noise, and clutter.

### 5.2. Floorplan Reconstruction

We implemented the proposed framework fully automatically, and tuned the parametric filtering threshold using a grid search approach on the validation set. The qualitative results of the proposed floorplan generation framework for a few sample buildings are provided in Figure 5.

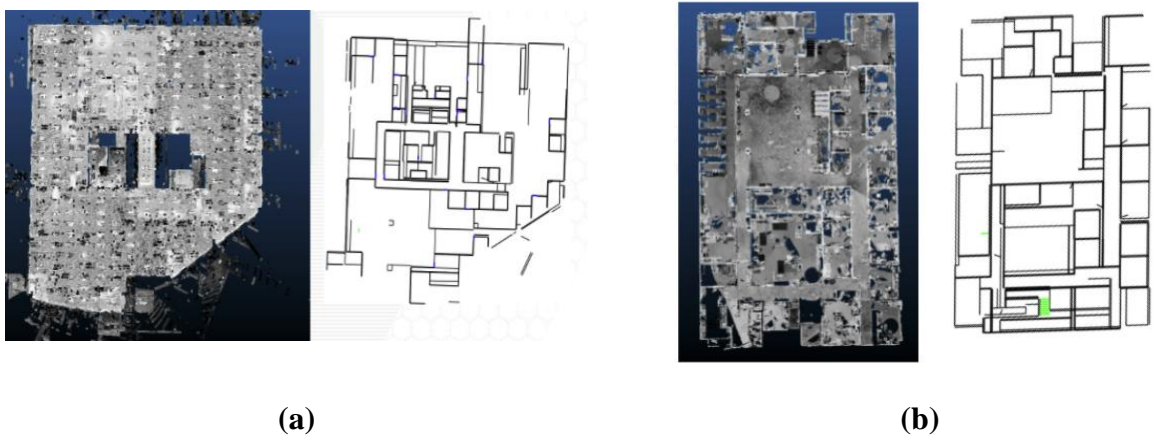

(a)                    (b)

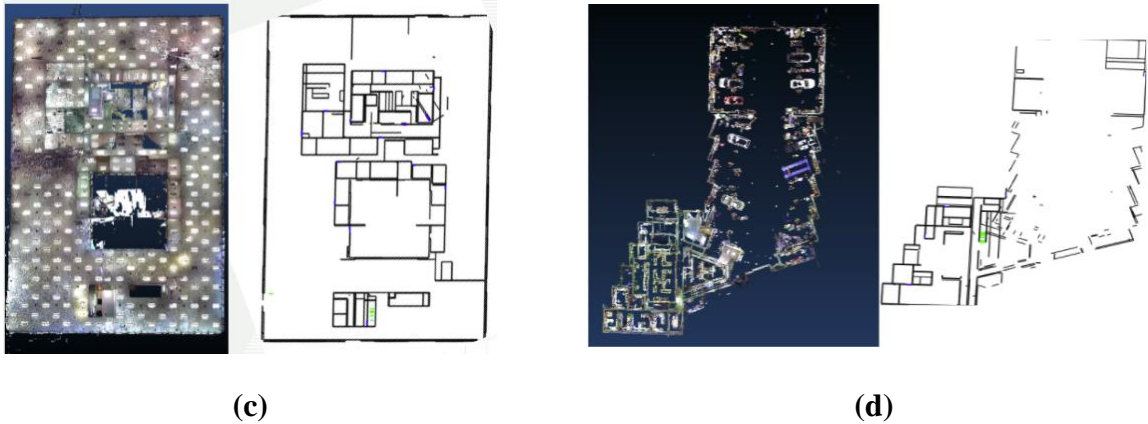

(c)　　　　　　　　　　　　　　　　　(d)

**Figure 5.** Floorplan reconstruction results

### 5.3. Evaluation

The evaluation results on the proposed framework (Table 3) can be found in *https://cv4aec.github.io/*. The evaluation metrics consist of both geometric and topological metrics. Geometric metrics include precision and recall at 2cm, 5cm, 10cm margins for each pixel in the generated floorplan. While for the topological metrics, each enclosed space is regarded as a room, and the IoU, Warping error, and Betti error are measured on the detected rooms. The proposed method ranks between 2nd and 3rd compared to other methods on geometric metrics. The proposed method achieves the best result on topological metrics such as warping error and Betti error.

**Table 3.** Qualitative evaluation of floorplan reconstruction

|      | Prec. (2cm) | Prec. (5cm) | Pre. (10cm) | Rec. (2cm) | Rec. (5cm) | Rec. (10cm) | IoU | Warping Error | Betti Error |
|---|---|---|---|---|---|---|---|---|---|
| [5] | 1.14% | 4.23% | 6.52% | 7.09% | 25.57% | 38.59% | 11.98% | 0.268 | 1.204 |
| [3] | 5.73% | 23.39% | 38.54% | 2.09% | 8.48% | 13.59% | 56.88% | 0.256 | 1.186 |
| **Ours** | 2.18% | 9.69% | 19.12% | 0.88% | 0.41% | 8.18% | 32.75% | **0.232** | **1.132** |

### 6. CONCLUSION

This research proposes a floorplan reconstruction framework for large-scale building point clouds and demonstrates that the hybrid approach of data-driven model and geometric reasoning can robustly improve the performance, especially in situations where the target building point clouds have diverse types and contain significant noise and clutter. Moreover, we implemented the framework in a fully automated manner, which is a significant advantage compared to many other methods for generating building models which require intermediate manual steps. The proposed automated framework paired with a pseudo-annotation method, being applicable even when the training data or model is incomplete, can serve as a stepping stone for increasing research interest in scanned building data and reconstruction of indoor environments.

There is much room for future research, as our results are still limited and the reconstruction accuracy is not sufficient for a real-world Scan-to-BIM application. For example, the class imbalance is an inevitable issue in point cloud processing because most of the scanned points come from wall, floor, or ceiling classes whereas door and stair classes are only a small minority in the scan data. In future work, we will approach this problem with advanced data augmentation methods and an improved loss function, thus enabling a better balance between the data-driven model and geometric reasoning.

## ACKNOWLEGEMENTS

This work is supported by the National Science Foundation (NSF) (Award #: OIA-2040735). Any opinions, findings, and conclusions or recommendations expressed in this material are those of the authors and do not necessarily reflect the views of NSF.